\documentclass[11pt,a4paper]{article}
\usepackage[hyperref]{acl2021}
\usepackage{times}
\usepackage{latexsym}

\usepackage{indentfirst}
\usepackage{graphicx}
\usepackage{subcaption}
\usepackage{multirow}
\usepackage{enumitem}

\usepackage{array, makecell}
\usepackage{microtype}
\usepackage{soul}

\aclfinalcopy 

\title{IITK@Detox at SemEval-2021 Task 5: Semi-Supervised Learning and Dice Loss for Toxic Spans Detection}

\author{Archit Bansal \qquad    
  Abhay Kaushik \qquad  
  \large{\textbf{Ashutosh Modi}} \\
{Indian Institute of Technology Kanpur (IIT Kanpur)} \\
  {\tt \{architb, kabhay\}@iitk.ac.in}  \\
  {\tt ashutoshm@cse.iitk.ac.in}  \\
}
\date{}

\begin{document}
\maketitle
\begin{abstract}
    In this work, we present our approach and findings for SemEval-2021 Task 5 - Toxic Spans Detection. The task's main aim was to identify spans to which a given text's toxicity could be attributed. The task is challenging mainly due to two constraints: the small training dataset and imbalanced class distribution. Our paper investigates two techniques, semi-supervised learning and learning with Self-Adjusting Dice Loss, for tackling these challenges. Our submitted system (ranked ninth on the leader board) consisted of an ensemble of various pre-trained Transformer Language Models trained using either of the above-proposed techniques.
\end{abstract}

\section{Introduction}
\noindent Content moderation has become the topic of most conversations regarding social media platforms. However, with over 4 billion active internet users, it is impossible to moderate each piece of message generated online manually. Therefore, the focus is now shifting towards tackling the issue using machine learning methods. 

Various toxicity detection datasets \cite{wulczyn2017ex, DBLP:journals/corr/abs-1903-04561} and models \cite{pavlopoulos-etal-2017-deeper, liu2019nuli, seganti-etal-2019-nlpr} have been successfully developed over the years to tackle the issue of moderation. However, these have mostly focused on identifying whole comments or documents as either toxic or not. In semi-automated settings, a model merely generating a toxicity score for each comment, some of which can be very lengthy, is not of much help to human moderators. To tackle this issue, the \textit{SemEval 2021 Task 5 : Toxic Spans Detection} is introduced \cite{pav2020semeval}. The task involves identifying text spans in a given toxic post that contributes towards the toxicity of that post. The task aims to promote the development of a system that would augment human moderators by giving them more insights into what actually contributes to the text's toxicity.

The task is challenging mainly due to the following reasons: a) small size of the dataset b) characteristics of text samples extracted from social media leading to difficulties such as out-of-vocabulary words and ungrammatical sentences c) class imbalance in the dataset d) inconsistencies in data annotations. We approached this task as a sub-token level sequence labeling task. Fine-tuned pre-trained transformer language models \cite{qiu2020pre} are the backbone of all our approaches. We investigated two main techniques to enhance the results of the fine-tuned transformer models, namely Semi-Supervised Learning \cite{yarowsky-1995-unsupervised, liu-etal-2011-recognizing} and fine-tuning with Self-Adjusting Dice Loss \cite{li2020dice}. This paper reports the results of our experiments with these different techniques and pre-trained transformer models.
Our submitted system consisted of an ensemble of different pre-trained transformer models and achieved an F1 score of 0.6895 on the test set and secured 9th position on the task leaderboard. All of our code is made publicly available on Github\footnote{\url{https://github.com/architb1703/Toxic_Span}}.

The rest of this paper is organized as follows. Section \ref{sec:Related} discusses the previous works in the fields of offensive language detection and span identification. Section \ref{ref:data} describes the dataset. Section \ref{sec:Approach} explains the proposed approaches. Section \ref{sec:res} reports the results of various experiments with the proposed approaches, and section \ref{sec:abalation} analyzes the proposed approaches via ablation studies. We conclude with an error analysis of our model performance in section \ref{sec:err} and concluding remarks in section \ref{sec:conc}.

\section{Related Work} \label{sec:Related}
\noindent As the task involves detecting toxic spans in a text, we present the related work in two parts: (i) Offensive Language Detection and (ii) Span Identification.

\textbf{Offensive Language Detection: }
Research work has been done on different abusive and offensive language identification problems, ranging from aggression \cite{kumar2018benchmarking} to hate
speech \cite{davidson2017automated}, toxic comments \cite{saif2018classification}, and offensive language \cite{laud2020problemconquero,pitsilis2018detecting}. Recent contributions to offensive language detection came from the SemEval-2019 Task 6 OffensEval \cite{zampieri2019semeval}. The task organizers concluded that most top-performing teams either used BERT \cite{liu2019nuli} or an ensemble model to achieve SOTA results. Interestingly, the task of locating toxic spans is relatively novel, and its successful completion can be groundbreaking. A recent approach with a narrower scope is by \citet{mathew2020hatexplain}, who focused on the rationality of decision in the task of hate speech detection.
\par
\textbf{Span Identification:}
Span detection/identification tasks include numerous tasks like named entity recognition (NER) \cite{nadeau2007survey}, chunking \cite{sang2000introduction} and keyphrase detection \cite{augenstein2017semeval}. \cite{papay2020dissecting} analyzed the span identification tasks via performance prediction over various neural architectures and showed that the presence of BERT component in the model is the highest positive predictor for these tasks. Inspired by this observation, we have built our model based on the transformer architecture, further exploiting the benefits of semi-supervised learning and modified Dice Loss. 
\section{Dataset}
\label{ref:data}

\subsection{Data Description}

\noindent The competition dataset comprises around 10K comments extracted from the Civil Comments Dataset and annotated using crowd-raters. The organizers released the dataset in 3 phases: trial, train, and test. The trial dataset consisted of 690 texts, whereas the training dataset consisted of 7939 texts. Moreover, the test set on which our system was finally evaluated consisted of 2000 text samples.
\par
In the initial stages of the competition, we decided to use only the training dataset to build upon our approaches. We further split the training set into train, dev, and test sets for evaluation purposes using an 80:10:10 split (Div A). Once we tested and finalized our approaches, we combined the train and test set of Div A with the trial set as our final training set (Div B).
Due to the small size of the dataset, these additions to the training set of Div A will positively impact the model performance. However, to ensure that we could compare our final models with our previous results, we transfer the dev set directly to Div B. Further details regarding the constitution of these splits is provided in the Appendix \ref{app:data}.
\subsection{Pre-processing}
\textbf{Tokenization:} 
For the sake of preserving the token spans, we first tokenized our data and then performed data cleaning. For tokenizing, we used the NLTK TreebankWord Tokenizer\footnote{\url{https://www.nltk.org/_modules/nltk/tokenize/treebank.html}}, which is a rule-based tokenizer that tokenizes text on spaces and punctuation, hence preserving the original form of the words.

\textbf{Data Cleaning:} We then cleaned each token using different operations such as expanding contractions and removing digits and full stops.  

\begin{figure*}[htp]
\begin{subfigure}{.44\textwidth}
    \centering
    \includegraphics[width=\linewidth]{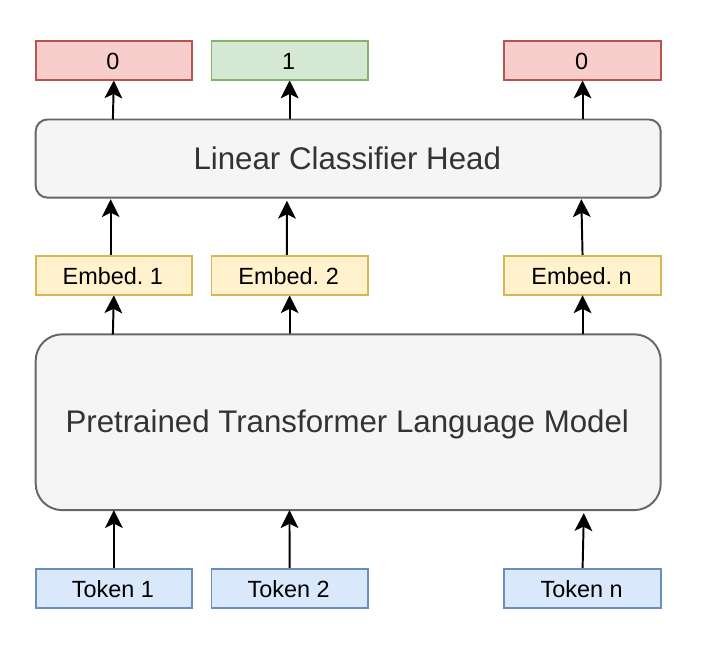}
    \subcaption{Sequence Labelling with Transformer Model}
    \label{fig:trans}
\end{subfigure}%
\begin{subfigure}{.52\textwidth}
    \centering
    \includegraphics[width=\linewidth]{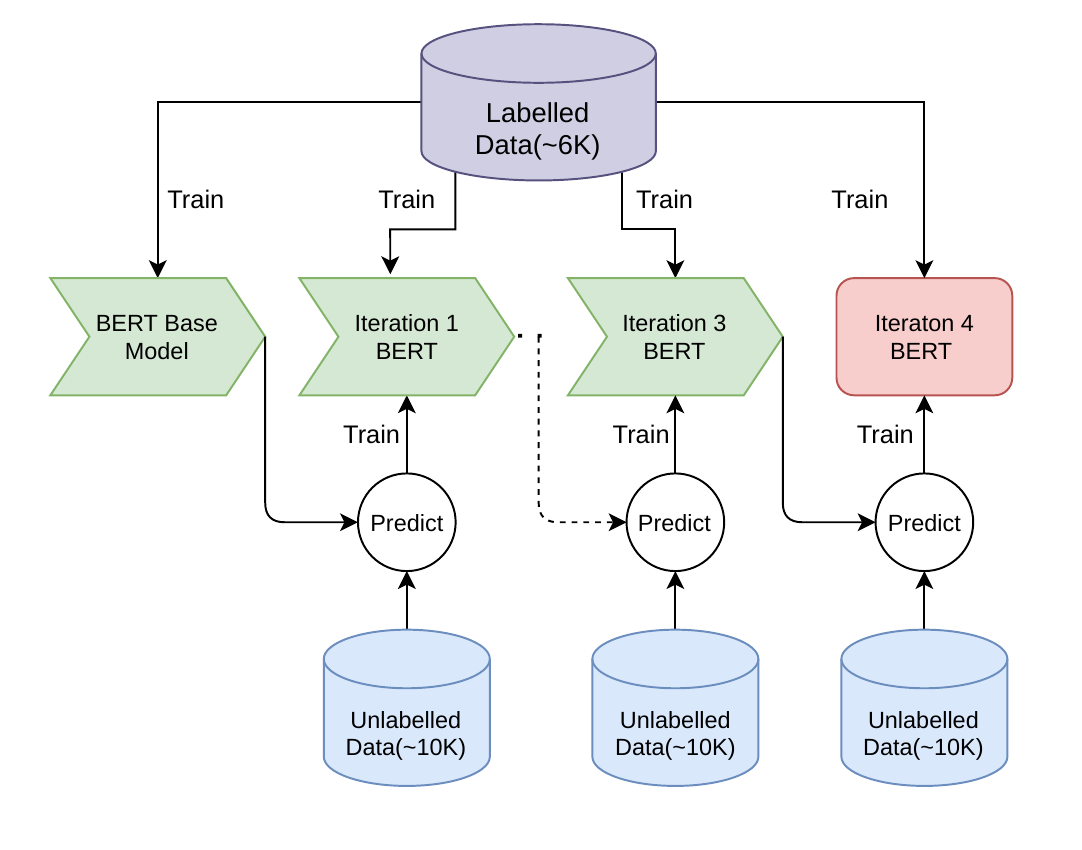}
    \subcaption{Semi-Supervised Learning architecture}
    \label{fig:SSL}
\end{subfigure}    
\caption{Model architectures}
\label{fig:test}
\end{figure*}

\section{Proposed Approach}
\label{sec:Approach}
\subsection{Methodology}
\noindent Pre-trained transformer models built using the transformer architecture \cite{vaswani2017attention} have been able to achieve, via transfer learning techniques, SOTA performance for most NLP tasks in recent times. We fine-tuned pre-trained transformer models with linear classifier head for performing sequence labeling for this task, which meant performing subtoken-level classification (Fig.\ref{fig:trans}). Our baseline model used the pre-trained BERT-Base-Cased model, fine-tuned with cross-entropy loss and AdamW optimizer. The different hyperparameter values used for training the baseline and all subsequent models are reported in the Appendix \ref{modtr} to facilitate replication of results. Subsequently, we improved upon this baseline using two techniques, semi-supervised learning and Self Adjusting Dice Loss. Along with this, we fine-tuned multiple different transformer models like BERT\cite{devlin2019bert}, Electra\cite{clark2020electra}, DistilBERT\cite{sanh2020distilbert}, and XLNet\cite{yang2020xlnet} for our Dice-Loss Approach and found differences in the predictions of the different transformer models to be beneficial for the final ensemble.

 \textbf{Self Adjusting Dice Loss}
 One of the main issues with our dataset was that of class imbalance. For the sub-tokens derived from the BERT-Base-Cased tokenizer, the ratio of toxic to non-toxic sub-tokens was 1:10.16. However, we could not tackle this issue with over/under-sampling due to the nature of our problem, and training with a weighted cross-entropy loss function did not improve results. Therefore, we experimented with training with the Self-Adjusting Dice Loss \cite{li2020dice} which was proposed as an objective function for dealing with imbalanced datasets in NLP. The original dice coefficient is an F1-oriented statistic used to gauge the similarity of two sets. The paper proposed a loss function based on a modified dice coefficient, which they reported to achieve a better F1 score than models trained with cross-entropy loss. 
 \[DL = 1 - \frac{2(1-p_{i1})^\alpha(p_{i1}).y_{i1} + \gamma}{(1-p_{i1})^\alpha(p_{i1}) + y_{i1} + \gamma}\]
 Here, for the $i_{th}$ training instance,  $p_{i1}$ is the predicted probability of positive class and $y_{i1}$ is the ground truth label. The loss function also has two hyperparameters, alpha and gamma, which we tuned for our models.
 
 \par\textbf{Semi-Supervised Learning}
The Civil Comments Dataset from which our training data was extracted consists of over 1 million comments; however, due to annotation constraints, the training set only had 7000 data samples. \cite{DBLP:journals/corr/Shams14a} have shown that for text classification tasks, unlabelled data from a suitable data source could be used to train semi-supervised models that achieve better results than a model trained using supervised learning. Also, \cite{jurkiewicz2020applicaai} showed that the semi-supervised learning technique of self-training could improve performance on span identification tasks. Hence, we extracted 40000 toxic samples from the Civil Comments Dataset, which were labeled with a toxicity score of 0.7 or higher, and used these to perform four iterations of semi-supervised model training (Fig. \ref{fig:SSL}). We exhaustively divided the unlabelled samples into four batches of 10000 each and used each batch for exactly one iteration. As shown in Fig. \ref{fig:SSL}, for each iteration, pseudo labels were predicted for the complete batch using the model trained in the previous iteration, then these pseudo-labels along with the ground truth training labels were used to train the next model. For this approach, we only fine-tuned one transformer model, namely the pre-trained BERT-Base-Cased model.

\subsection{Post-preprocessing}
\noindent After obtaining the sub-token level labels from our model, we post-processed the results to convert them into an array of toxic character offsets. To perform this, we had mapped each sub-token to its offset span during tokenization and used that to retrieve the offsets of all the characters in the toxic sub-tokens. We also include all characters lying between two consecutive sub-tokens if both the sub-tokens are marked toxic. This was necessary as spaces and punctuation were included in the toxic spans given by the annotators, as shown by the results in section \ref{sec:abalation}.

\section{Experiments and Results}
\label{sec:res}
\noindent The competition used the span level F1 score, calculated individually for each text sample from its character offsets and averaged over all the text samples, as the metric to evaluate system performance. We decided to use this metric for hyperparameter tuning and reporting the final results. However, during the training process, the models were checked for overfitting using the token level F1 score, which was also a good indicator of model performance as our training approach was that of a sequence labeling task.

The first set of experiments we performed were on the Div-A dataset. Our baseline model achieved an F1 score of 0.669 on the dev split on this set. The organizers also released a baseline model, consisting of a Spacy statistical model trained on the competition training dataset and evaluated on the competition trial dataset. The organizer's baseline achieved an F1 score of 0.600 on the trial dataset. This score was significantly lower than that of our baseline model, and therefore we use our baseline model only to compare the performance of our subsequent models.  
\par
We then fine-tuned a BERT-Base-Cased model with the Self-Adjusting Dice Loss and AdamW optimizer and tuned the loss function's two hyperparameters. The scores we obtained for the different hyperparameter values are reported in Table \ref{tab:dicepara} in Appendix \ref{sec:rst}. We got our best performing model with the hyperparameter values alpha-0.7 and gamma-0.25, achieving an F1 score of 0.6725 on the dev split.

\begin{table}[!hbt]
    \centering
    \begin{tabular}{|c|c|c|}
    \hline
    Model & Dev F1 Score\\ 
    \hline
    BERT-Base-Cased & 0.669  \\
    SSL Iteration-1 & 0.6837\\
    SSL Iteration-2 & 0.6842 \\
    SSL Iteration-3 & 0.6882 \\
    SSl Iteration-4 & \textbf{0.6893} \\
    \hline
    \end{tabular}
    \caption{Results for Semi-Supervised learning model}
    \label{tab:SSL_start}
\end{table}
\par
Next, we trained the BERT-Base-Cased model on the semi-supervised learning paradigm with cross-entropy loss and AdamW optimizer. For the first iteration, we used our baseline model to compute the pseudo labels. The model achieved improved results with each iteration (Table \ref{tab:SSL_start}), and our final model was scoring 0.6893 on the dev split.

To end this stage of experimentation, we computed the results on the test split of the Div-A dataset. We were able to make two inferences. Firstly the semi-supervised learning model had the best performance with an F1 score of 0.6774 on the test set but had a significantly worse score than it had on the dev set. Secondly, the dice loss trained model performed significantly better than the cross-entropy trained baseline with an F1 score of 0.662 compared to 0.648. 

After this, we changed to the Div-B dataset and trained multiple different transformer models with the Self Adjusting Dice Loss. We found that the BERT-Base-Cased, Electra-Small, Electra-Base, and DistilBert-Base-Uncased models had peak performance for the hyperparameter values alpha-0.7 and gamma-0.25.  However, for the XLNet-Base model, peak performance was achieved for alpha-0.4 and gamma-0.25. On further experimentation with these models, we also found that adding a full stop to the text samples during evaluation provided consistently better results on the dev set. The results obtained have been reported in Table \ref{tab:preproces}. 

\begin{table}[!hbt]
    \centering
    \begin{tabular}{|c|c|c|}
    \hline
    Model & WFS & FS \\ 
    \hline
     BERT-Base-Cased & 0.6754 & 0.6827\\
     Electra-Small & 0.6813 & 0.6861\\
     Electra-Base & 0.6776 & 0.6846\\
     DistilBERT-Base-Unc. & 0.6749 & 0.6773\\
     XLNet-Base & 0.6798 & 0.6852\\
     SSl Iteration-4 & 0.6893 & 0.6932\\
     \hline
    \end{tabular}
    \caption{Effect of full stop on dev set during evaluation. Here WFS and FS represent without full stop and with full stop resp.}
    \label{tab:preproces}
\end{table}

The final results of models trained either on modified Dice Loss or using Semi-Supervised learning, with full stop added during evaluation, are reported on the Div-B dev split and the competition test set in Table \ref{tab:SSL}.

\begin{table}[!hbt]
    \centering
    \begin{tabular}{|c|c|c|c|}
    \hline
    & Model & Dev & Test\\ 
    \hline
    1 & BERT-Base-Cased* & 0.6827 & 0.668\\
    2 & Electra-Small* & 0.6861 & 0.6771\\
    3 & Electra-Base* & 0.6846 & 0.6720\\
    4 & DistilBERT-Base-Unc.* & 0.6773 & 0.6822\\
    5 & XLNet-Base* & 0.6852 & 0.6757\\
    6 & SSl Iteration-4 & 0.6932 & 0.672\\
     & Ensemble (1,2,3,4,5,6) & 0.6927 & \textbf{0.6895}\\
    \hline
    \end{tabular}
    \caption{F1 Score on dev and competition test set \newline  
    * - Models trained with modified Dice Loss}
    \label{tab:SSL}
    \vspace{-0.4cm}
\end{table}

\section{Ablation Study}
\label{sec:abalation}
\noindent After the competition, we wanted to study the effect of our different preprocessing and postprocessing techniques. We employ three main data cleaning techniques during our preprocessing, expanding contractions, removing numbers, and removing full stops. To study each particular technique's impact, we created three new Div-B datasets, each having one of the preprocessing techniques missing. We then trained BERT-Base-Cased and Electra-Small models with Self Adjusting Dice Loss on each of these sets and evaluated the performance on their respective dev sets. The results are reported in Table \ref{tab:abl} with the following acronyms:

\begin{itemize}[noitemsep]
    \item \textbf{TD} - All preprocessing steps used
    \item \textbf{WNUM} - Without removing numbers 
    \item \textbf{WFS} - Without removing fullstops
    \item \textbf{WCON} - Without expanding contractions
\end{itemize}

\begin{table}[!hbt]
    \centering
    \begin{tabular}{|c|c|c|}
    \hline
    Dataset & BERT-Base-Cased & Electra-Small\\ 
    \hline
    TD & 0.6754 & 0.6813\\
    WNUM & \textbf{0.6781} & 0.6809\\
    WFS & 0.6713 & 0.6743\\
    WCON & 0.671 & \textbf{0.6829}\\
    \hline
    \end{tabular}
    \caption{F1 Score for different preprocessing techniques on dev set}
    \label{tab:abl}
\end{table}
The results show that removing numbers and expanding contractions both had contrasting effects on the two models. This shows that we could have yielded better results by trying different preprocessing techniques for the different transformer models. Apart from that, we see that the most positive effect on model performance came from removing full stops from the training data in both cases.

We also wanted to see the effect of our postprocessing step. For that, we compared the performance of the BERT-Base-Cased model on the Div-B dev split. As expected, the results showed minor improvement due to our postprocessing as the score increased from 0.6748 to 0.6754.

\section{Error Analysis}
\label{sec:err}
\noindent The results we have obtained have brought to light some problems that need to be resolved. First of all, the data annotations have many issues, leading to a lower F1 score even though the predicted toxic spans are more appropriate in many cases. We have included some examples in Appendix \ref{app:err}. In some cases, the annotations are not uniform in what toxicity label they assign to the same word over different text samples. We have also observed that complete sentences were marked as toxic just because of the presence of a few toxic words in them. These irregularities in the annotations make it difficult for the model to generalize on the data.

Besides the incorrect annotations, we further try to analyze the type of mistakes our system is making. 
The dataset contains numerous examples where no toxic spans are annotated. Such a case arose when the annotators had difficulty in attributing toxicity to a particular span. Investigating our model performance shows that our model highly under-performs on such examples. Table \ref{tab:err} depicts the drastic difference in the performance of the system over empty span examples (E.S) and non-empty span examples (N.E.S). Upon closely following E.S examples, we discovered that annotations of such examples carry more subjectivity than the others. In such cases, our model usually labels the word with the most negative sentiment as toxic and thus performs poorly. 

 \begin{table}[]
    \centering
    \begin{tabular}{|c|c|c|}
    \hline
        Example Set & Val & Test\\ 
        \hline
        \makecell{ E.S \\ Val:41 Test:394} & 0.0731 & 0.0380 \\
        \hline
        \makecell{ N.E.S \\Val:753 Test:1606)}&0.7265 &0.8493\\
        \hline
        \makecell{All \\ Val:794 Test :2000}& 0.6927 & 0.6895\\
        \hline
    \end{tabular}
    \caption{System performance over empty span (E.S) and non-empty span(N.E.S) examples over Div-B split}
    \label{tab:err}
\end{table}

In addition to the empty span examples, we also discover that our model fails to capture the full context in some cases. For e.g., in the phrase ``no more Chinese," our model only predicts the word Chinese as toxic, whereas the complete phrase attributes to the toxicity of the sentence. Another problem is our model's inconsistency in labeling the corresponding noun and adjective pairs in a sentence. However, similar types of inconsistencies were also found in the annotations and are therefore difficult to avoid [Appendix \ref{app:err}].

\section{Conclusion}
\label{sec:conc}
\noindent The task of detecting toxic spans in the text is a novel one, and there is no doubt about how important a model trained successfully for this task can turn out to be for online content moderation. However, the data gathered from online platforms tend to be noisy and corrupted. Coupled with the limitations of generating large-scale annotated datasets in real life, they pose two daunting challenges. In conclusion, our final submission shows that transfer learning through pre-trained transformer models can achieve competitive results for this task. Using modified loss functions and semi-supervised learning, even more can be extracted from limited annotated data. Moreover, considering the subjectivity involved in span detection, the task can also be expanded to report severity scores of spans and classify the type of toxicity. This will further help simplify and rationalize online content moderation.

\bibliographystyle{acl_natbib}
\bibliography{acl2021}

\begin{thebibliography}{28}
\expandafter\ifx\csname natexlab\endcsname\relax\def\natexlab#1{#1}\fi

\bibitem[{Augenstein et~al.(2017)Augenstein, Das, Riedel, Vikraman, and
  McCallum}]{augenstein2017semeval}
Isabelle Augenstein, Mrinal Das, Sebastian Riedel, Lakshmi Vikraman, and Andrew
  McCallum. 2017.
\newblock Semeval 2017 task 10: Scienceie-extracting keyphrases and relations
  from scientific publications.
\newblock \emph{arXiv preprint arXiv:1704.02853}.

\bibitem[{Borkan et~al.(2019)Borkan, Dixon, Sorensen, Thain, and
  Vasserman}]{DBLP:journals/corr/abs-1903-04561}
Daniel Borkan, Lucas Dixon, Jeffrey Sorensen, Nithum Thain, and Lucy Vasserman.
  2019.
\newblock \href {http://arxiv.org/abs/1903.04561} {Nuanced metrics for
  measuring unintended bias with real data for text classification}.
\newblock \emph{CoRR}, abs/1903.04561.

\bibitem[{Clark et~al.(2020)Clark, Luong, Le, and Manning}]{clark2020electra}
Kevin Clark, Minh-Thang Luong, Quoc~V. Le, and Christopher~D. Manning. 2020.
\newblock \href {http://arxiv.org/abs/2003.10555} {Electra: Pre-training text
  encoders as discriminators rather than generators}.

\bibitem[{Davidson et~al.(2017)Davidson, Warmsley, Macy, and
  Weber}]{davidson2017automated}
Thomas Davidson, Dana Warmsley, Michael Macy, and Ingmar Weber. 2017.
\newblock Automated hate speech detection and the problem of offensive
  language.
\newblock \emph{arXiv preprint arXiv:1703.04009}.

\bibitem[{Devlin et~al.(2019)Devlin, Chang, Lee, and
  Toutanova}]{devlin2019bert}
Jacob Devlin, Ming-Wei Chang, Kenton Lee, and Kristina Toutanova. 2019.
\newblock \href {http://arxiv.org/abs/1810.04805} {Bert: Pre-training of deep
  bidirectional transformers for language understanding}.

\bibitem[{Jurkiewicz et~al.(2020)Jurkiewicz, Borchmann, Kosmala, and
  Grali{\'n}ski}]{jurkiewicz2020applicaai}
Dawid Jurkiewicz, {\L}ukasz Borchmann, Izabela Kosmala, and Filip
  Grali{\'n}ski. 2020.
\newblock Applicaai at semeval-2020 task 11: On roberta-crf, span cls and
  whether self-training helps them.
\newblock \emph{arXiv preprint arXiv:2005.07934}.

\bibitem[{Kumar et~al.(2018)Kumar, Ojha, Malmasi, and
  Zampieri}]{kumar2018benchmarking}
Ritesh Kumar, Atul~Kr Ojha, Shervin Malmasi, and Marcos Zampieri. 2018.
\newblock Benchmarking aggression identification in social media.
\newblock In \emph{Proceedings of the First Workshop on Trolling, Aggression
  and Cyberbullying (TRAC-2018)}, pages 1--11.

\bibitem[{Laud et~al.(2020)Laud, Singh, Sahu, and
  Modi}]{laud2020problemconquero}
Karishma Laud, Jagriti Singh, Randeep~Kumar Sahu, and Ashutosh Modi. 2020.
\newblock problemconquero at semeval-2020 task 12: Transformer and soft
  label-based approaches.
\newblock In \emph{Proceedings of the Fourteenth Workshop on Semantic
  Evaluation}, pages 2123--2132.

\bibitem[{Li et~al.(2020)Li, Sun, Meng, Liang, Wu, and Li}]{li2020dice}
Xiaoya Li, Xiaofei Sun, Yuxian Meng, Junjun Liang, Fei Wu, and Jiwei Li. 2020.
\newblock \href {http://arxiv.org/abs/1911.02855} {Dice loss for
  data-imbalanced nlp tasks}.

\bibitem[{Liu et~al.(2019)Liu, Li, and Zou}]{liu2019nuli}
Ping Liu, Wen Li, and Liang Zou. 2019.
\newblock Nuli at semeval-2019 task 6: Transfer learning for offensive language
  detection using bidirectional transformers.
\newblock In \emph{Proceedings of the 13th International Workshop on Semantic
  Evaluation}, pages 87--91.

\bibitem[{Liu et~al.(2011)Liu, Zhang, Wei, and
  Zhou}]{liu-etal-2011-recognizing}
Xiaohua Liu, Shaodian Zhang, Furu Wei, and Ming Zhou. 2011.
\newblock \href {https://www.aclweb.org/anthology/P11-1037} {Recognizing named
  entities in tweets}.
\newblock In \emph{Proceedings of the 49th Annual Meeting of the Association
  for Computational Linguistics: Human Language Technologies}, pages 359--367,
  Portland, Oregon, USA. Association for Computational Linguistics.

\bibitem[{Mathew et~al.(2020)Mathew, Saha, Yimam, Biemann, Goyal, and
  Mukherjee}]{mathew2020hatexplain}
Binny Mathew, Punyajoy Saha, Seid~Muhie Yimam, Chris Biemann, Pawan Goyal, and
  Animesh Mukherjee. 2020.
\newblock Hatexplain: A benchmark dataset for explainable hate speech
  detection.
\newblock \emph{arXiv preprint arXiv:2012.10289}.

\bibitem[{Nadeau and Sekine(2007)}]{nadeau2007survey}
David Nadeau and Satoshi Sekine. 2007.
\newblock A survey of named entity recognition and classification.
\newblock \emph{Lingvisticae Investigationes}, 30(1):3--26.

\bibitem[{Papay et~al.(2020)Papay, Klinger, and Pad{\'o}}]{papay2020dissecting}
Sean Papay, Roman Klinger, and Sebastian Pad{\'o}. 2020.
\newblock Dissecting span identification tasks with performance prediction.
\newblock \emph{arXiv preprint arXiv:2010.02587}.

\bibitem[{Pavlopoulos et~al.(2021)Pavlopoulos, Laugier, Sorensen, and
  Androutsopoulos}]{pav2020semeval}
John Pavlopoulos, Léo Laugier, Jeffrey Sorensen, and Ion Androutsopoulos.
  2021.
\newblock Semeval-2021 task 5: Toxic spans detection (to appear).
\newblock In \emph{Proceedings of the 15th International Workshop on Semantic
  Evaluation}.

\bibitem[{Pavlopoulos et~al.(2017)Pavlopoulos, Malakasiotis, and
  Androutsopoulos}]{pavlopoulos-etal-2017-deeper}
John Pavlopoulos, Prodromos Malakasiotis, and Ion Androutsopoulos. 2017.
\newblock \href {https://doi.org/10.18653/v1/D17-1117} {Deeper attention to
  abusive user content moderation}.
\newblock In \emph{Proceedings of the 2017 Conference on Empirical Methods in
  Natural Language Processing}, pages 1125--1135, Copenhagen, Denmark.
  Association for Computational Linguistics.

\bibitem[{Pitsilis et~al.(2018)Pitsilis, Ramampiaro, and
  Langseth}]{pitsilis2018detecting}
Georgios~K Pitsilis, Heri Ramampiaro, and Helge Langseth. 2018.
\newblock Detecting offensive language in tweets using deep learning.
\newblock \emph{arXiv preprint arXiv:1801.04433}.

\bibitem[{Qiu et~al.(2020)Qiu, Sun, Xu, Shao, Dai, and Huang}]{qiu2020pre}
Xipeng Qiu, Tianxiang Sun, Yige Xu, Yunfan Shao, Ning Dai, and Xuanjing Huang.
  2020.
\newblock Pre-trained models for natural language processing: A survey.
\newblock \emph{Science China Technological Sciences}, pages 1--26.

\bibitem[{Saif et~al.(2018)Saif, Medvedev, Medvedev, and
  Atanasova}]{saif2018classification}
Mujahed~A Saif, Alexander~N Medvedev, Maxim~A Medvedev, and Todorka Atanasova.
  2018.
\newblock Classification of online toxic comments using the logistic regression
  and neural networks models.
\newblock In \emph{AIP conference proceedings}, volume 2048, page 060011. AIP
  Publishing LLC.

\bibitem[{Sang and Buchholz(2000)}]{sang2000introduction}
Erik~F Sang and Sabine Buchholz. 2000.
\newblock Introduction to the conll-2000 shared task: Chunking.
\newblock \emph{arXiv preprint cs/0009008}.

\bibitem[{Sanh et~al.(2020)Sanh, Debut, Chaumond, and
  Wolf}]{sanh2020distilbert}
Victor Sanh, Lysandre Debut, Julien Chaumond, and Thomas Wolf. 2020.
\newblock \href {http://arxiv.org/abs/1910.01108} {Distilbert, a distilled
  version of bert: smaller, faster, cheaper and lighter}.

\bibitem[{Seganti et~al.(2019)Seganti, Sobol, Orlova, Kim, Staniszewski,
  Krumholc, and Koziel}]{seganti-etal-2019-nlpr}
Alessandro Seganti, Helena Sobol, Iryna Orlova, Hannam Kim, Jakub Staniszewski,
  Tymoteusz Krumholc, and Krystian Koziel. 2019.
\newblock \href {https://doi.org/10.18653/v1/S19-2126} {{NLPR}@{SRPOL} at
  {S}em{E}val-2019 task 6 and task 5: Linguistically enhanced deep learning
  offensive sentence classifier}.
\newblock In \emph{Proceedings of the 13th International Workshop on Semantic
  Evaluation}, pages 712--721, Minneapolis, Minnesota, USA. Association for
  Computational Linguistics.

\bibitem[{Shams(2014)}]{DBLP:journals/corr/Shams14a}
Rushdi Shams. 2014.
\newblock \href {http://arxiv.org/abs/1409.7612} {Semi-supervised
  classification for natural language processing}.
\newblock \emph{CoRR}, abs/1409.7612.

\bibitem[{Vaswani et~al.(2017)Vaswani, Shazeer, Parmar, Uszkoreit, Jones,
  Gomez, Kaiser, and Polosukhin}]{vaswani2017attention}
Ashish Vaswani, Noam Shazeer, Niki Parmar, Jakob Uszkoreit, Llion Jones,
  Aidan~N. Gomez, Lukasz Kaiser, and Illia Polosukhin. 2017.
\newblock \href {http://arxiv.org/abs/1706.03762} {Attention is all you need}.

\bibitem[{Wulczyn et~al.(2017)Wulczyn, Thain, and Dixon}]{wulczyn2017ex}
Ellery Wulczyn, Nithum Thain, and Lucas Dixon. 2017.
\newblock \href {http://arxiv.org/abs/1610.08914} {Ex machina: Personal attacks
  seen at scale}.

\bibitem[{Yang et~al.(2020)Yang, Dai, Yang, Carbonell, Salakhutdinov, and
  Le}]{yang2020xlnet}
Zhilin Yang, Zihang Dai, Yiming Yang, Jaime Carbonell, Ruslan Salakhutdinov,
  and Quoc~V. Le. 2020.
\newblock \href {http://arxiv.org/abs/1906.08237} {Xlnet: Generalized
  autoregressive pretraining for language understanding}.

\bibitem[{Yarowsky(1995)}]{yarowsky-1995-unsupervised}
David Yarowsky. 1995.
\newblock \href {https://doi.org/10.3115/981658.981684} {Unsupervised word
  sense disambiguation rivaling supervised methods}.
\newblock In \emph{33rd Annual Meeting of the Association for Computational
  Linguistics}, pages 189--196, Cambridge, Massachusetts, USA. Association for
  Computational Linguistics.

\bibitem[{Zampieri et~al.(2019)Zampieri, Malmasi, Nakov, Rosenthal, Farra, and
  Kumar}]{zampieri2019semeval}
Marcos Zampieri, Shervin Malmasi, Preslav Nakov, Sara Rosenthal, Noura Farra,
  and Ritesh Kumar. 2019.
\newblock Semeval-2019 task 6: Identifying and categorizing offensive language
  in social media (offenseval).
\newblock \emph{arXiv preprint arXiv:1903.08983}.

\end{thebibliography}

\clearpage
\section*{Appendix}
\appendix
\section{Dataset}
\label{app:data}
We worked on two different splits of the data across different stages of competition. Table \ref{tab:data} represent the no. of examples in train, val and test across Div-A and Div-B split.
\begin{table}[!hbt]
    \centering
    \begin{tabular}{|c|c|c|}
    \hline
         & Div-A & Div-B\\ 
    \hline
        Train & 6351 & 7835\\ 
        Dev & 794 & 794\\
        Test & 794 & 2000\\
        Total & 7939 & 10629\\
    \hline
    \end{tabular}
    \caption{Distribution of examples across Div-A and Div-B split}
    \label{tab:data}
\end{table}
\par
Div-A is basically a 80:10:10 split of the training data released by the organisers whereas Div-B split uses the train and test set of Div-A along with competition trial data as its training set. Div-B uses the official test set as its test set while keeping the dev set same as that of Div-A.

\section{Model Training}
\label{modtr}
In this section, we provide the hyperparameter values we used while training our final models to facilitate the replication of our results at a later time. The acronyms correspond to:
\begin{itemize}[noitemsep]
    \item \textbf{LR} : Learning Rate
    \item \textbf{ML} : Max Len
    \item \textbf{LC} : Data Lowercase
    \item \textbf{DL} : Dice Loss (Alpha, Gamma)
\end{itemize}

\begin{table}[!hbt]
    \centering
    \begin{tabular}{|c|c|c|}
    \hline
        Hyperpara. & BERT-Base-Cased & Electra\\ 
    \hline
        LR & 1E-5 & 3E-5\\
        ML & 500 & 500\\
        LC & False & True\\
        DL & 0.7,0.25 & 0.7,0.25\\
    \hline
    \end{tabular}
    \caption{Hyperparameter Values for BERT-Base-Cased and Electra (Small and Base)}
    \label{tab:dice1}
\end{table}

\begin{table}[!hbt]
    \centering
    \begin{tabular}{|c|c|c|}
    \hline
        Hyperpara. & Distil-Base-Unc. & XLNet-Base\\ 
    \hline
        LR & 1E-5 & 3E-5\\
        ML & 500 & 400\\
        LC & True & False\\
        DL & 0.7,0.25 & 0.4,0.25\\
    \hline
    \end{tabular}
    \caption{Hyperparameter Values for Distil-Base-Unc. and XLNet-Base}
    \label{tab:dice2}
\end{table}

\noindent For baseline model and semi-supervised learning model, the cross-entropy loss function provided in PyTorch was used with default hyperparameters. For AdamW optimizer, we used weight decay rate of 0.01 for all model parameters apart from the bias, gamma and beta terms, for which we used 0 weight decay. 

\section{Results}
\label{sec:rst}
\subsection{Dice Loss Results}
In this section, we report the results of our hyperparameter tuning for training with Self Adjusting Dice Loss. The result reported are for BERT-Base-Cased model on the Div-A dataset.
\begin{table}[!hbt]
    \centering
    \begin{tabular}{|c|c|}
    \hline
        Parameter Values & Dev F1 Score\\ 
    \hline
        Alpha-0, Gamma-1 & 0.665\\ 
        Alpha-0, Gamma-2 & 0.657\\
        Alpha-0.5 Gamma-1 & 0.652\\
        Alpha-0.7, Gamma-1 & 0.671\\
        Alpha-0.8, Gamma-1 & 0.668\\
        Alpha-0.7, Gamma-0.5 & 0.6723\\
        Alpha-0.7, Gamma-0.25 & \textbf{0.6725}\\
    \hline
    \end{tabular}
    \caption{BERT-Base-Cased trained with Self Adjusting Dice Loss on Div-A train split}
    \label{tab:dicepara}
\end{table}

\section{Error Analysis}
\label{app:err}
We manually crawled through a part of dataset and found several inadequate annotations as described
through table \ref{tab:ErrAnalysis}.
Apart from erroneous annotations, we also found some examples of inconsistency in the structure of annotations. Table \ref{tab:err2} outlines a few such pairs where different strategies are used for annotating similar texts. These examples highlight the inconsistencies in labelling of corresponding noun-adjective pairs.  
\begin{table}[hbt]
    \centering
    \begin{tabular}{|c|c|}
        \hline
         Text & Ground Truth \\
         \hline
         You are an idiot&['idiot']\\
         You are truly witless & All tokens\\
         \hline
         F***ing moron .... &['F***ing']\\
         Fucking moron needs... & ['Fucking','moron']\\
         \hline
         Hilary is a liar... & All tokens\\
         Trump is an idiot. & ['idiot']\\
         \hline
    \end{tabular}
    \caption{Irregularity in annotation for similar text-samples.}
    \label{tab:err2}
\end{table}

\begin{table*}[htp]
    \begin{tabular}{|c|c|c|}
    \hline
        Text & Ground Truth  & Suggested\\ 
    \hline
        \makecell{I have no doubt it was Photoshopped(by an idiot, for an idiot)} & ['idiot'] & ['idiot','idiot'] \\
        \hline
        She's a Marxist twit. &	['Marxist']	
         & ['Marxist', 'twit' ]\\
        \hline 
         \makecell{Move to Russia and be a fascist pig all you want.} &	All tokens &	['fascist', 'pig'] \\
         \hline
         \makecell{lame. really really lame and cheap shot. pathetic actually.}& \makecell{['lame', 'lame', \\'actually']}& \makecell{['lame', 'lame',\\ 'pathetic']}\\
         \hline
         \makecell{So "f*** you all", "kiss my a--" and ...}& ['f***'] & ['f***','a--']\\
         \hline
         \makecell{You are just a stick buthole!}&[] &['stick','buthole']\\
    \hline
    \end{tabular}
    \caption{Shows the words labelled toxic as ground truth(from annotators) and suggested accurate annotations }
    \label{tab:ErrAnalysis}
\end{table*}

\end{document}